\documentstyle[12pt]{article}
\input epsf
\begin{document}
\renewcommand{\baselinestretch}{0.9}
\small \normalsize
\newcommand{\datr}{{\sc DATR}}
\newcommand{\ling}{\it}
\newtheorem{guess}{Fig.}

\begin{center} {\Large\bf Automatic Inference of \datr\ Theories  }\\[0.5cm]
  Petra Barg$^*$
 \\[0.3cm]
{\ Seminar f\"ur Allgemeine Sprachwissenschaft, Universit\"at D\"usseldorf,
Universit\"atsstra{\ss}e 1, D-40225 D\"usseldorf, Germany}\\
\end{center} \vskip0.5cm
{\small \bf Summary:} {\ An approach for the automatic acquisition of
linguistic knowledge from unstructured data is presented. The acquired
knowledge is represented in the lexical knowledge representation
language \datr. A set of transformation rules that establish inheritance
relationships and a default-inference algorithm make up the basis components
of the system. Since the overall approach is not restricted to a special
domain, the heuristic inference strategy uses criteria to evaluate
the quality of a DATR theory, where different domains may require
different criteria. The system is applied to the linguistic learning task
of German noun inflection.}\\[20pt]
{\large \bf 1. Introduction }\\[10pt]
The following paper presents an approach for automatic acquisition
of linguistic knowledge from observations within a given domain;
it thus addresses a topic from the field of machine learning
(cf. Michalski (1986)), or more precisely, from the field of machine learning
of natural language, as it is sometimes called
(cf. Powers and Reeker (1991)).\\[5pt]
The last decade has seen a growing interest in the application of
machine learning to different kinds of linguistic domains
(cf. Powers and Reeker (1991)) and has been motivated by different objectives,
such as modelling cognitive processes or overcoming the knowledge-acquisition
bottleneck in natural-language processing systems. The principal motivation
for our approach is theoretical. The automatically induced analyses can
be compared to existing linguistic descriptions and thus can confirm
proposed analyses or provide alternative representations. On the other
hand, the approach can be used by descriptive linguists as a tool to
obtain a rough structuring of an entirely new domain (e.g. a language
that has not yet been investigated).\\[5pt]
Theoretical
approaches and implemented systems cover subjects from many different
linguistic areas and use different kinds of learning strategies. Some are
specially designed for a particular task, while others are more general and
can thus be applied to other tasks as well. Whereas the present paper
pursues a general approach that is not restricted to a specific
linguistic domain, the system is crucially determined by the properties
of the chosen representation language.
\\[5mm]
\footnoterule
{\footnotesize \hspace{4mm} $^*$ The presented work was partly supported by the
Deutsche Forschungsgemeinschaft (DFG).
For helpful comments and discussions on the topic
we would like to thank Gerald Gazdar, Dafydd Gibbon, James Kilbury, Ingrid
Renz, and Markus Walther.}
\newpage
In designing a learning system the choice of a language for representing
the acquired knowledge is crucial for the quality of the output of the
learning task. One requirement for a linguistic representation formalism
is that the information can be structured in a way that captures
generalizations over linguistic objects in order to minimize redundancy.
Since many linguistic generalizations have exceptions that can only be
treated adequately if the representation formalism includes some device for
handling default information, we have chosen the language \datr\
(cf. Evans and Gazdar (1989), (1990)), which allows regularities and
sub-/irregularities to be expressed in a uniform way. \\[5pt]
We briefly summarize the main features of \datr\ here but presuppose a basic
familiarity with the language as described in (Evans and Gazdar (1989)).
\datr\ is a declarative formalism for the definition of inheritance networks.
It includes orthogonal multiple inheritance and a
default-inheritance mechanism.
A network description in \datr\ is called a {\it theory}
and describes a set of objects ({\it nodes}). The properties of an object
are defined by {\it path-definition} pairs, where a path consists of an
ordered sequence of atoms (enclosed in angle brackets). The definition can
be either the directly stated {\it value} (atomic value or sequence
of atomic values) of the property, an {\it inheritance descriptor} that
states where the value of that property can be inherited from, or a
{\it sequence} of inheritance descriptors. An inheritance descriptor can
refer to another node, path or node-path pair of the theory. The triple
consisting of a node, a path, and a definition is called a
{\it definitional sentence}.\\[5pt]
The simple \datr\ theory in Fig.~\ref{theory} encodes
information about English verb morphology. It contains the three
node definitions {\tt VERB}, {\tt Love}, and {\tt Come}, which contain three,
two, and four definitional sentences, respectively. The node definition
{\tt VERB} encodes the information that all past tense forms of a verb
are like the root plus {\it \_ed}, and all present tense forms are like the
root, with the exception of the form for third singular. As a regular verb
{\tt Love} inherits all information except the morphological root from
the node {\tt VERB}. In contrast {\tt Come} deviates from the regular
verbs in its past tense forms, which are therefore
specified in the node definition. All other information can be inherited
from {\tt VERB}.
\begin{guess}
\label{theory}
a simple \datr\ theory
\end{guess}
{\tt
\begin{tabbing}
\hspace{1.2cm}
\= VERB: \= $<$mor past$>$ == ("$<$mor root$>$" \_ed)\\
     \>  \> $<$mor pres tense$>$ == "$<$mor root$>$"\\
     \>  \> $<$mor pres tense sing three$>$ == ("$<$mor root$>$" \_s).\\

\> Love: \> $<$$>$ == VERB\\
   \>   \> $<$mor root$>$ == love.\\

\> Come: \> $<$$>$ == VERB\\
   \>   \> $<$mor root$>$ == come\\
   \>   \> $<$mor past$>$ == came\\
   \>   \> $<$mor past participle$>$ == $<$mor root$>$.
\end{tabbing}
}
The information expressed in a \datr\ theory is accessed by {\it queries}
concerning objects and their properties. A query consists of a
node-path pair and returns an atomic value (or a sequence of atomic values)
or fails. Seven inference rules and a default mechanism are given to
deterministically evaluate the queries.
The query {\tt Love:$<$mor pres tense sing two$>$} evaluates to {\tt love}
for the theory in Fig.~\ref{theory}. A query together with its
returned value is called an {\it extensional sentence}.\vspace{15pt}

{\large \bf 2. Inference of \datr\ theories }\\[10pt]
Many learning systems use the same formal language to represent the input data
and the acquired knowledge. Extensional sentences (which constitute the
output of the conventional inference in \datr) form a natural sublanguage
of \datr\ which is suitable to represent the input data. Since extensional
sentences all have atomic values and thus are not related to each
other, they can be taken as representing independent and unstructured facts
about a given linguistic domain. The learning task then consists in
forming a \datr\ theory which accounts for the observed facts through
adequate structuring.\footnote{Light (1994) addresses a related topic,
the insertion of a new object (described with extensional \datr\
sentences) into an existing \datr\ theory. In contrast to our approach
the assumption of a structured initial theory is made.}\\[5pt]
For an acquired \datr\ theory to be regarded as adequately characterizing
a given set of observations it has to meet at least the following criteria
(in addition to the general syntactic wellformedness conditions that hold
for every \datr\ theory):
\begin{itemize}
\item  consistency with respect to the input data
\item  completeness with respect to the input data
\item  structuring of the observed data by inheritance relationships
\item  structuring of the observed data by generalizing them
\end{itemize}
The first two of these criteria constitute minimal, formal requirements
that can be verified easily. A \datr\ theory is consistent with respect to
a given set of extensional sentences if, for every query that constitutes
the left-hand side of one of the extensional sentences, the returned value
is that of the extensional sentence. If this holds for all left-hand sides
of the extensional sentences the theory is also complete with respect to
the input data.\\[5pt]
The last two criteria rely more on intuitions and cannot be checked so easily.
The inferred \datr\ theory should structure the observed data so that it
reveals relationships that exist between the extensional sentences. A \datr\
theory expresses such relationships by the use of inheritance descriptors.\\[5pt]
The generalization of the observed data is twofold. First of all, a set of
specific facts should be generalized, whenever this is possible, to a
single more general assumption that covers all of the specific facts.
In \datr\, such generalizations are captured by defaults expressed in sentences
that cover more than one property of an object (as opposed to the input data,
where each sentence is supposed to represent a single observed property).
For example, the sentence {\tt VERB:$<$mor past$>$ == ("$<$mor root$>$" \_ed)}
of the theory in Fig.~\ref{theory} covers all past tense forms of a verb.
In addition to this process of generalization which is used in many
machine-learning systems (e.g. Mitchell (1982), Michalski (1983)), acquired
\datr\ theories should identify information that several objects have in common.
This information should be abstracted and stored in more general objects
from which the others inherit. Such generalized objects further structure
the domain because hierarchies evolve where objects are grouped into classes.
\\[10pt]
{\bf 2.1 Acquisition of inheritance relationships }\\[5pt]
The observed data constitute a trivial \datr\ theory which forms the initial
hypothesis $H_0$ of the learning task. This \datr\ theory is complete and
consistent with respect to the input
but does not meet the other two criteria. This section addresses
the question of how a given \datr\ theory can be transformed into another
theory that contains more inheritance descriptors or changes the
latter in order to structure the domain.\\[5pt]
The knowledge of how a given \datr\ theory can be transformed into a new one with
different inheritance descriptors is defined by rewrite rules of the
following format:
\begin{guess}
\label{transrule}
form of a transformation rule
\end{guess}
\hspace{1.2cm} $s_i \rightarrow {s_i}' / c_1,..,c_n$\\[5pt]
where $s_i$ is the input sentence and ${s_i}'$ is the transformed sentence. Since
inheritance descriptors are stated as right-hand sides (RHS) or parts of
RHSs of sentences, the transformation rules operate on RHSs of \datr\ sentences.
Thus, ${s_i}'$ differs from $s_i$ in that it contains a different RHS.
$c_1,..,c_n$ are constraints that define under what conditions a given
sentence can be transformed into another one. In order to carry out a
transformation that maintains the completeness and consistency of the theory
a major constraint for the application of most transformation rules
to a hypothesis $H_i$ consists in the requirement that $H_i$ contain
another sentence with the same RHS as the sentence that
is to be transformed.\\[5pt]
Corresponding to the different kinds of inheritance relationships that
can be expressed in a \datr\ theory, there are four major groups of
transformation rules: rules that return sentences with local descriptors
(local paths, local nodes, local node-path pairs), rules that transform
sentences into others that have a global descriptor, rules where the
transformed sentence contains a descriptor that refers to a sentence with
a global descriptor, and rules that create new, abstract sentences for
the acquisition of a hierarchy.\footnote{Barg (1995) gives a full account
of all transformation rules.}
In Fig.~\ref{localnode}
the rule for creating local node
descriptors is formulated. Here, $H_i$ is the given \datr\ theory,
$V_a$ is the set of atomic values in $H_i$, and $N$ is the set of nodes in
$H_i$. The rule transforms a sentence $s$ with atomic value into one with
a node descriptor $v'$, if the theory contains another sentence $s_i$
that belongs to node $v'$ and has the same path and value as $s$.
\begin{guess}
\label{localnode}
rule for local node inheritance
\end{guess}
\begin{eqnarray*}
\hspace{-2cm}
s:(n,p,v) \rightarrow s^\prime:(n,p,v^\prime) & / & v \in V_a,   \\
                                              &   & v^\prime \in N, \\
                                              &   & s_i:(v^\prime,p,v) \in H_i, \\
                                              &   & s_i \neq s
\end{eqnarray*}
By means of transformation rules all the different kinds of inheritance
descriptors can be obtained with the exception of evaluable paths.
Evaluable paths capture dependencies between properties with
different values and therefore cannot be acquired by
transformation rules that crucially depend on the existence of sentences
which have the same RHSs. Therefore, they have here been excluded from
the learning task.\\[10pt]
{\bf 2.2 Acquisition of default information}\\[5pt]
While inheritance relationships are represented with the RHSs of sentences,
default information is basically expressed through paths of the left-hand
sides (LHSs), namely by paths that cover more than one fact. Since
transformation rules leave the LHSs of sentences unchanged, an additional
device is necessary that operates on LHSs of sentences. For this purpose a
default-inference algorithm (DIA) was developed that reduces any given \datr\
theory that does not (yet) contain default information, where ''reduction''
means shortening the paths of sentences (by cutting off a path suffix) or
deletion of whole sentences. Since extensive generalization is normally
a desirable property, the resulting theory must be (and indeed is)
maximally reduced.\\[5pt]
In order to acquire a \datr\ default theory that remains consistent with
respect to the input data
the DIA has to check that a reduction of a sentence
does not lead to any conflicts with the remaining sentences of the theory.
Conflicts can only arise between sentences which have the same node and path,
because in all other cases the longest matching path can be determined.
Therefore, if a given sentence is to be shortened, it has to be checked
whether the theory already contains another sentence with the same node
and shortened path. If it does, and if the other sentence has a different
RHS, the first sentence cannot be shortened and must remain in the resulting
theory. If the other sentence has the same RHS, the first sentence can
be removed from the theory altogether. If the theory does not contain the
shortened sentence, the shortening is a legitimate operation since no
conflicts can arise.\\[5pt]
The following additional restrictions must be imposed to guarantee a
theory that is complete and consistent with respect to the input data.
First of all, the sentences of a node have
to be considered in descending order according to the length of their paths.
This guarantees that for every sentence, the sentences it can conflict with
are still contained in the theory and are not shortened or removed.
For similar reasons, sentences can only be shortened by one element (the last)
at a time. In the case of path references or node-path pairs, some additional
tests are carried out since potential conflicts arise from \datr's
mechanism governing the inheritance of path extensions.
\vspace{15pt}

\begin{figure}
\begin{guess}
\label{hierarchy}
acquired hierarchy for German noun inflection
\end{guess}
\epsfbox[110 -79 669 724]{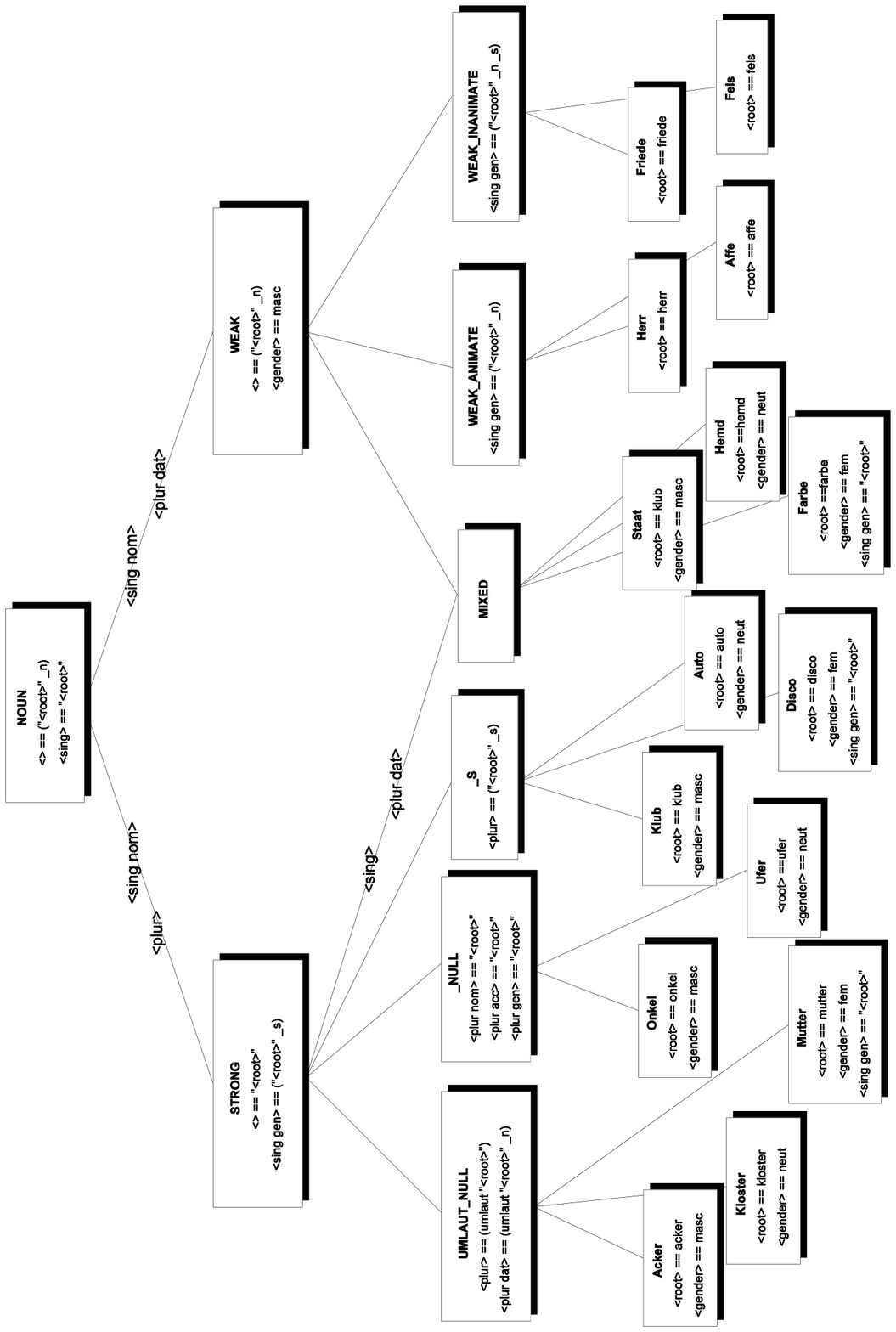}
\end{figure}

{\bf 2.3 Inference strategy }\\[5pt]
The inference strategy determines how a result hypothesis $H_R$ is acquired
from an initial hypothesis $H_0$. It relies on the notion of a permissible
derivation which arises through applications of transformation rules and DIA.
A permissible derivation of $H_0$ results from any sequence of transformation
rules followed by the DIA. For reasons of consistency it is not possible
to apply transformation rules after the DIA or to apply the DIA several times.
\\[5pt]
Many different theories can be derived from $H_0$, but only some of them can be
regarded as good \datr\ theories with respect to the input data.\footnote{The
question of what constitutes a good \datr\ theory is addressed later.
Assume for the moment that it has been defined and that two \datr\
theories can be compared with each other with respect to quality.}
In order to acquire a good theory the space of permissible derivations
has to be searched. Since an exhaustive search leads to a combinatorial
explosion for every non-trivial problem, a heuristic search is used as
in many other systems. We employ a forward pruning strategy that works
as follows: First of all, by further restricting the transformation rules
and DIA, not all of the possible successor hypotheses
are generated for a given hypothesis. Most importantly, the rules
for building hierarchies are
restricted in order to gain sensible classes. Here the notion of similarity
of objects (i.e. the number of sentences that two objects have in common)
plays a crucial role as in clustering approaches
(cf. Stepp and Michalski (1986), Lebowitz (1987)).\\[5pt]
Of the generated successor hypotheses only the few most promising ones
are further expanded, while all others are discarded from the search.
To decide which hypotheses are promising, criteria are needed to evaluate \datr\
theories. Since only monotonic \datr\ theories can be further transformed, these
criteria have to be formulated for such theories. On the other hand,
only default theories are considered as possible solutions, since the
representation of default information constitutes a major demand on an
appropriate theory. Therefore, the default theories resulting from the
most promising monotonic theories are the candidates for the result
hypothesis. Again, criteria are needed in order to select the best of these
candidates. The search terminates when no more transformation rules can be
applied.\footnote{This presupposes that the search space is finite,
which is guaranteed by further restricting the transformation rules
(more precisely the rules for creating abstract sentences).}\\[5pt]
Each kind of criteria forms a complex that is composed of various different
single criteria that are ordered according to priority. As the inference
strategy is not restricted to any specific domain, different learning tasks
usually require different evaluation criteria or different orderings.
Among the criteria that were found to be most useful are the following:
\begin{itemize}
\item size of a \datr\ theory, measured by the absolute or average number of sentences
per object (useful only for default theories)
\item homogeneity of RHSs, measured by the number of different RHSs
\item complexity of RHSs (length of paths and sequences)
\item capturing of particular relationships such as
\begin{itemize}
\item relationships between objects (relative number of node references)
\item relationships within objects (relative number of
path references)\vspace{15pt}
\end{itemize}
\end{itemize}
{\large \bf 3. Inference of German noun inflection }\\[10pt]
An implementation of the approach has been applied to a number of different
learning tasks, including the acquisition of German noun inflection
(cf. Wurzel (1970)). The input for these tasks can be drawn from sample
evaluations of a corresponding \datr\ theory that is included in the \datr\
papers (cf. Evans and Gazdar (1990)). It consists of sentences whose paths contain
attributes for case and number and whose values are the inflected word forms
(here, abstract morphemes) associated with them, as illustrated in
Fig.~\ref{input}.
In addition, information about the root form and the gender are included.
\begin{guess}
\label{input}
input sentence for German noun inflection
\end{guess}
\hspace{1.3cm}
{\tt Fels: <plur~nom> = (fels~{\_}n).}\\[5pt]
For the learning task observations about nouns of various inflectional
classes are given: {\ling Fels} 'rock', {\ling Friede} 'peace',
{\ling Herr} 'gentleman' and {\ling Affe} 'monkey' are weak nouns,
{\ling Staat} 'state', {\ling Hemd} 'shirt' and {\ling Farbe} 'colour'
are mixed, and {\ling Acker} 'field', {\ling Kloster} 'convent', {\ling Mutter}
'mother', {\ling Onkel} 'uncle', {\ling Ufer} 'shore', {\ling Klub} 'club',
{\ling Auto} 'car', and {\ling Disco} 'disco' are strong.\\[5pt]
The criteria for selecting the most promising theories during search were the
number of different references, followed by the complexity of inheritance
descriptors and number of levels in the hierarchy. The
criteria for determining the best hypotheses were the number of sentences with
a node-path pair on the RHS, the relative number of sentences with
no node reference, and the average size of objects. All of the mentioned
criteria were to be minimized.
The acquired \datr\ theory is depicted graphically in Fig.~\ref{hierarchy}.
Here, the automatically generated abstract node names are replaced
(manually) by more linguistically motivated names. Edges that are not
annotated correspond to inheritance via the empty path.\\[5pt]
The inferred hierarchy in Fig.~\ref{hierarchy} structures the domain of
German noun inflection in a linguistically plausible way.
According to similarity nouns are grouped into six major classes, from
which they inherit most of their information.
The first three of them (UMLAUT{\_}NULL, NULL, {\_}S) correspond to strong
classes that have in common the formation of singular forms but differ in
their plural forms, which are therefore stated explicitly.
The last two classes (WEAK{\_}ANIMATE, WEAK{\_}INANIMATE)
represent weak noun classes that differ only in the formation of their
forms for genitive singular. The commonalities of strong nouns on the one
hand and weak nouns on the other hand are further abstracted from these
classes and specified in the two more general node definitions
STRONG and WEAK respectively. As an interesting fact,
the class of mixed nouns (MIXED)
has been identified, whose members behave like strong nouns
with respect to the formation
of their singular forms and like weak nouns in the formation of their plural
forms. These facts are captured by inheriting information
from the classes STRONG and WEAK.
Finally, the top node NOUN of the hierarchy
represents information that is typical for German nouns in general.\vspace{15pt}

{\large \bf 4. Conclusion }\\[10pt]
This paper has presented an approach to the acquisition of linguistic
knowledge from unstructured data.
The approach is general in the sense that it is not restricted
to a specific linguistic domain. This has been achieved by choosing the
general representation language DATR for the representation of the acquired
knowledge and by postulating a learning strategy that is tailor-made for
this formalism. A similar approach could be conceived for other
knowledge-representation formalisms (e.g. KL-ONE,
cf. Brachman and Schmolze (1985))
which are more familiar within the artificial-intelligence paradigm.\\[5pt]
The system was applied to a learning task involving German noun inflection.
The results are sensible in that nouns are grouped into classes according
to their inflectional behavior in such a way that generalizations are captured.
The acquired theories are restricted in that they do not make use of evaluable
paths; thus, although they are clearly non-trivial, the theories constitute
a proper sublanguage of \datr. In the future, further applications of the system
within different domains must be made in order to get a more detailed view of
its possibilities. This pertains especially to the criteria for guiding
the search and selecting best hypotheses.\vspace{15pt}

{\bf References:}\\[10pt] {\small
BARG, P. (1995):
Automatischer Erwerb von linguistischem Wissen: ein Ansatz zur Inferenz
von DATR-Theorien. Dissertation, Heinrich-Heine-Universit\"at
D\"ussel\-dorf.\\[5pt]
BRACHMAN, R.J., and SCHMOLZE, J.G. (1985):
An Overview of the KL-ONE Knowledge Representation System. {\it Cognitive
Science}, 9, 171-216.\newpage
EVANS, R., and GAZDAR, G. (1989):
Inference in DATR. {\it Proc. of the 4th Conference of the European Chapter
of the Association for Computational Linguistics}, 66-71.\\[5pt]
EVANS, R., and GAZDAR, G. (eds.) (1990):
{\it The DATR Papers: February 1990 (= Cognitive Science Research Paper 139)}.
School of Cognitive and Computing Sciences, Univerity of Sussex,
Brighton, England.\\[5pt]
LEBOWITZ, M. (1987): Experiments with Incremental Concept Formation:
UNI\-MEM.
{\it Machine Learning}, 2, 103-138.\\[5pt]
LIGHT, M. (1994):
Classification in Feature-based Default Inheritance Hierarchies.
In H. Trost (ed.): {\it KONVENS '94: Verarbeitung nat\"urlicher Sprache}.
\"Osterreichische Gesellschaft f\"ur Artificial Intelligence, Wien,
220-229.\\[5pt]
MICHALSKI, R. (1983):
A Theory and Methodology of Inductive Learning.
{\it Artificial Intelligence}, vol. 20, 2, 111-161.\\[5pt]
MICHALSKI, R. (1986):
Understanding the nature of learning: Issues and research directions.
In R.S. Michalski, J.G. Carbonell, and T.M. Mitchell (eds.):
{\it Machine Learning: An Artificial Intelligence Approach}.
Los Altos, Morgan Kaufmann, vol. 2, 3-25.\\[5pt]
MITCHELL, T.M. (1982):
Generalization as search. {\it Artificial Intelligence}, vol. 18, 203-226.
\\[5pt]
POWERS, D. and REEKER, L. (1991):
{\it Machine Learning of Natural Language and Ontology (Proc. AAAI
Spring Symposium)}. Kaiserslautern.\\[5pt]
STEPP, R.E. and MICHALSKI, R.S. (1986):
Conceptual Clustering: Inventing Goal-Oriented Classifications of
Structured Objects. In R.S. Michalski, J.G. Carbonell, and T.M. Mitchell
(eds.): {\it Machine Learning: An Artificial Intelligence Approach}.
Los Altos, Morgan Kaufmann, vol. 2, 471-498.\\[5pt]
WURZEL, W. (1970):
{\it Studien zur deutschen Lautstruktur}. Akademie-Verlag, Berlin.

\end{document}